\documentclass[journal]{IEEEtran}
\ifCLASSINFOpdf
\usepackage[pdftex]{graphicx}
\usepackage{subfigure}
\else
\fi
\usepackage{amsmath}
\usepackage{amsfonts}
\usepackage{bbm}

\usepackage{algorithmic}
\usepackage{algorithm}
\usepackage{url}

\usepackage{multirow}
\usepackage{booktabs}

\begin{document}
%
\title{Interval Type-2 Fuzzy Neural Networks for Multi-Label Classification}

\author{Dayong Tian,
        Feifei Li,
        Yiwen Wei
\thanks{This work was supported in part by Natural Science Foundation of Shaanxi Province under Grant 2020JQ-197.}
\thanks{Dayong Tian and Feifei Li are with the School of Electronics and Information, Northwestern Polytechnical University, Xi'an, 710071, P.R. China.}
\thanks{Yiwen Wei is with the School of Physics and Optoelectronic Engineering, Xidian University, Xi'an, 71000, P.R. China.}
}

\maketitle

\begin{abstract}
Prediction of multi-dimensional labels plays an important role in machine learning problems. We found that the classical binary labels could not reflect the contents and their relationships in an instance. Hence, we propose a multi-label classification model based on interval type-2 fuzzy logic. In the proposed model, we use a deep neural network to predict the type-1 fuzzy membership of an instance and another one to predict the fuzzifiers of the membership to generate interval type-2 fuzzy memberships. We also propose a loss function to measure the similarities between binary labels in datasets and interval type-2 fuzzy memberships generated by our model. The experiments validate that our approach outperforms baselines on multi-label classification benchmarks.
\end{abstract}


%
\IEEEpeerreviewmaketitle

\section{Introduction}
\begin{figure}[h]
\centering
\includegraphics[width=0.9\linewidth, trim=80 0 80 0]{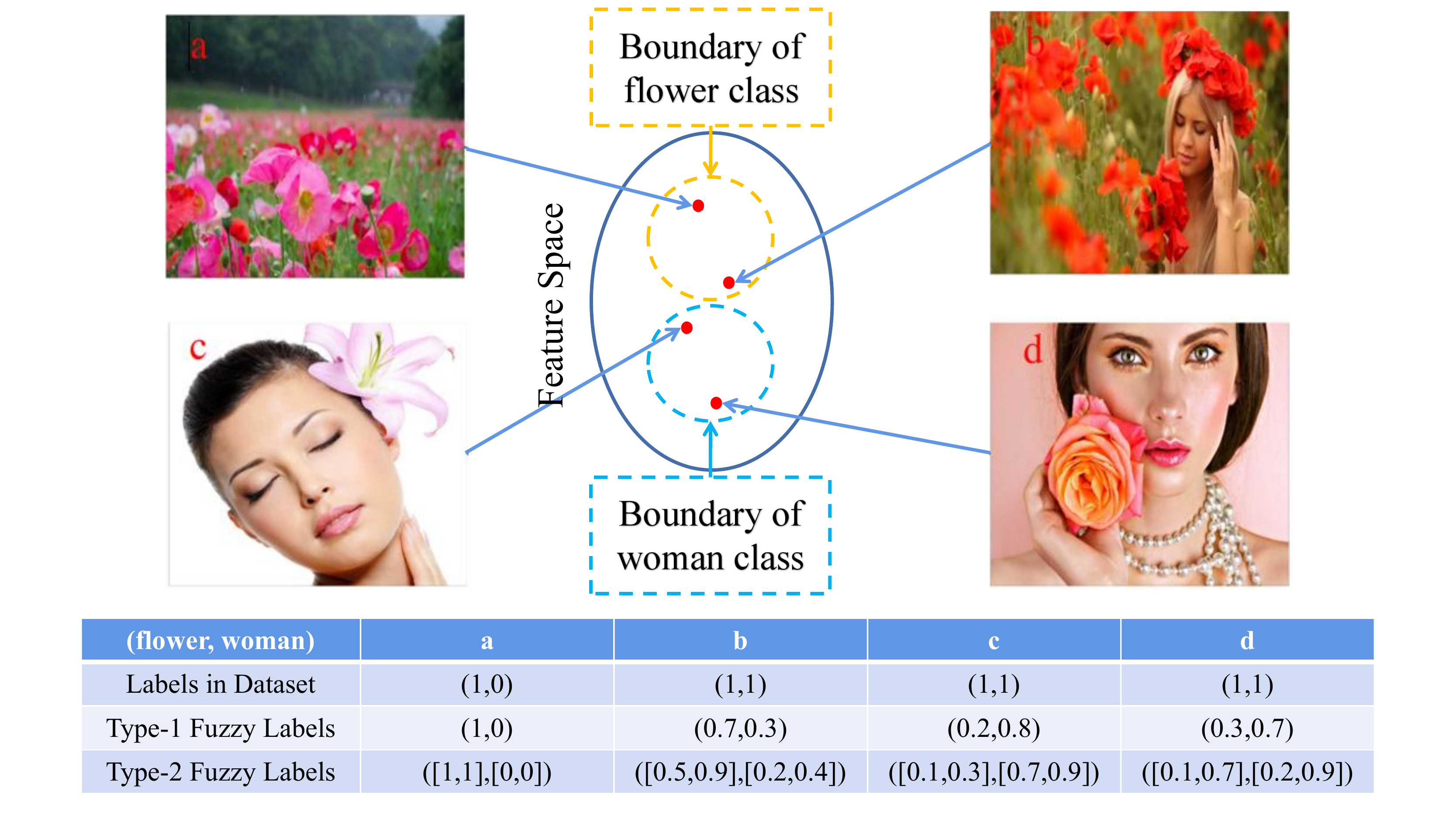}
\caption{Illustration of different types of labels. The labels in dataset are binary. Type-1 fuzzy labels may be generated from the judgment of a human expert. Type-2 fuzzy labels may be generated from the several human experts' judgment. To visualize the feature extraction of a learning model, say convolutional neural network, we assume the extracted features are two-dimensional. The images' features are plotted in the feature space.}
\label{fig:motivation}
\end{figure}

%
%
%
%
\IEEEPARstart{M}ulti-label classification (MLC) aims at predicting multiple labels by exploring the dependencies among labels. It has been widely used in natural language processing~\cite{chang_2019} and computer vision~\cite{wang_cnn-rnn_2016}.\\
\indent Classical MLC models make binary predictions. That is, they use 0 and 1 to indicate the categorizations. However, intuitively, human has fuzzy discrimination on an instance(Fig.~\ref{fig:motivation}).\\
\indent When categorizing these four images in Fig.~\ref{fig:motivation}, human observers may consider the semantic relationships between flowers or women in the images and hence they may give fuzzy estimations of their categories. For image (a), it apparently belongs to flower. For (b), someone may consider the woman is the main content while others may consider woman and flower are of the same importance. For (c), we human can make sure that the image belongs to the woman category and the flower is just a decoration. However, the flower occupies a considerably large area in the image. A MLC model may also consider that it belongs to flower category in certain degree. For (d), human observers' opinions may be contradictory again because someone may consider the woman holds the flower in front of her face so she probably want to show a special species of flower, while someone may consider that the flower is just an ordinary one and the main content of the image is still the woman.\\ 
\indent Almost in all datasets, instances are directly labeled by 0 or 1, which cannot effectively represent humans' fuzzy judgment. Interval type-2 fuzzy logic is capable of representing different human opinions on categorization. It is the motivation that we explore how to build a fuzzy-logic-based model for MLC. \\
\indent In uni-label classification, probability and type-1 fuzzy logic are a little bit confusing. Softmax is widely used in neural networks for uni-label classification. The output of softmax can be explained as the probability of an instance belonging to a category. Someone can ``illegally'' explain the probabilities as type-1 fuzzy memberships. If we are working on uni-label classification, there is few numerical difference between these two explanations. For example, we can explain Fig.~\ref{fig:motivation} as this. The probability of the image (a) belonging to flower category is maximum, so the image is belonging to flower category. The membership of the image (b) belonging to woman category is maximum. If we must make a decision on which category it belongs to, we have to categorize the image into woman category.\\
\begin{figure*}[th]
\centering
\includegraphics[width=0.7\linewidth]{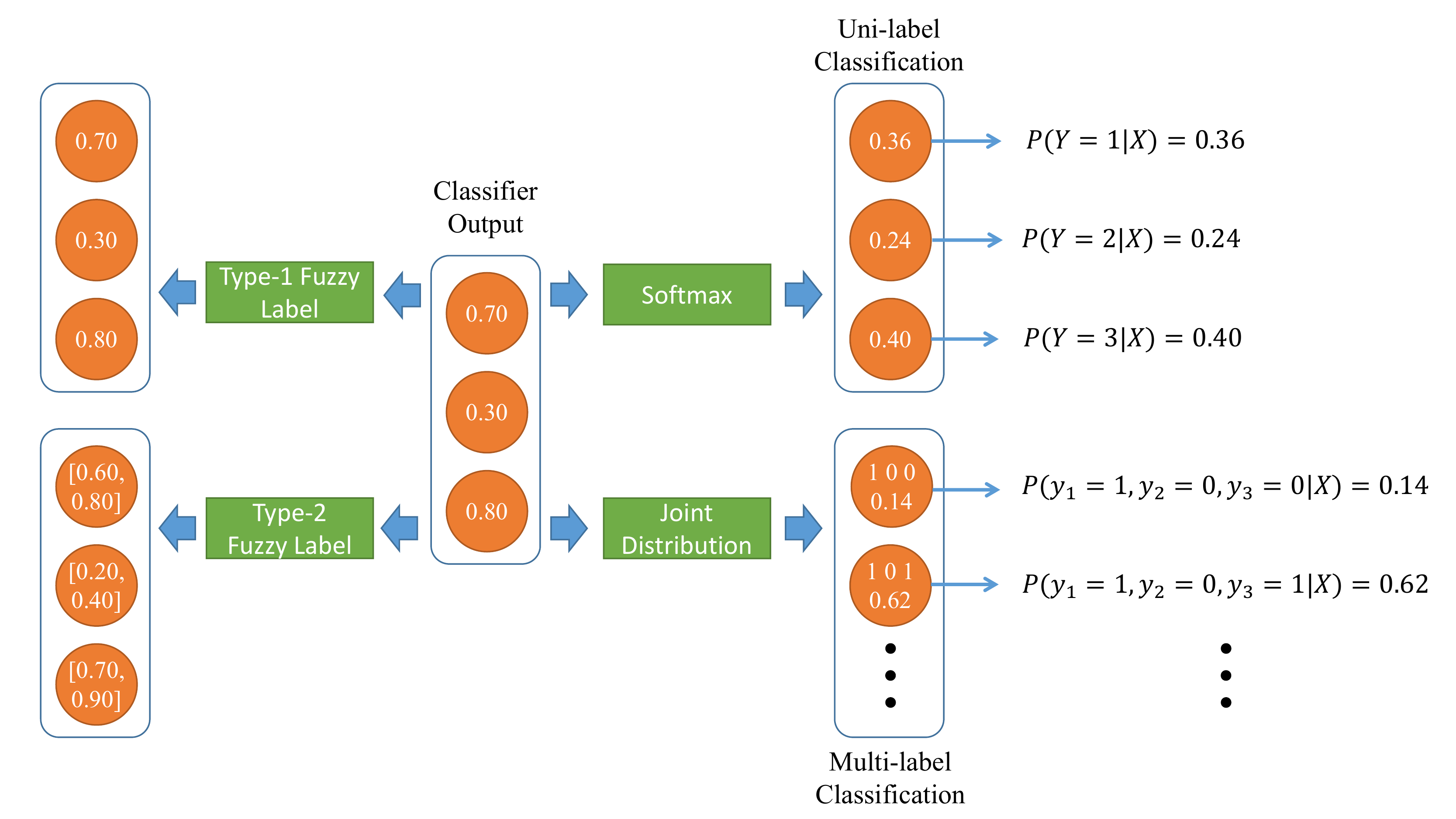}
\caption{The differences among probabilities, type-1 fuzzy logic and interval type-2 fuzzy logic. Suppose we have $L$ categories. In the multi-label classification case, we examine the joint probabilities of all elements in the \emph{label powerset} which is defined as a set of all possible label combinations. Hence, there are $2^L$ label combinations. If we directly explain the output of the classifier as type-1 or type-2 fuzzy predictions, we only need to examine $L$ fuzzy memberships.} 
\label{fig:fuzzy2}
\end{figure*}
\begin{figure*}[th]
\centering
\includegraphics[width=0.9\linewidth, trim=0 50 0 0]{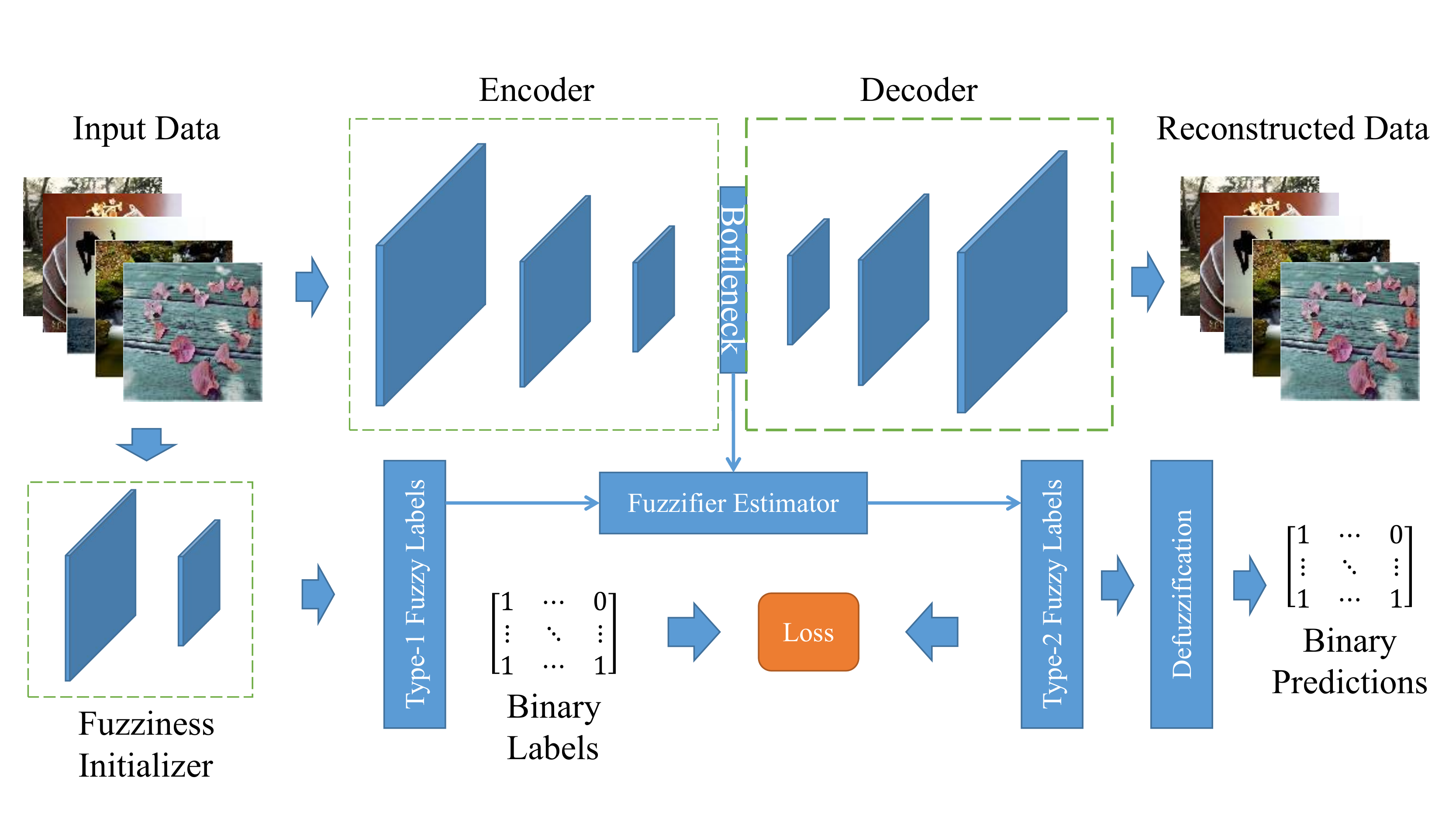}
\caption{The scheme of our method. The input data are used to train two neural networks, a fuzziness intializer and a fuzzifier estimator. The fuzziness intializer is used to generate type-1 fuzzy labels. The output of the bottleneck of the autoencoder is used as an input of a fuzzifier estimator. The estimated fuzzifiers are incorporated into type-1 fuzzy labels to generate interval type-2 fuzzy labels. Then, we design a loss function to measure the dissimilarities between binary labels in original dataset and generated interval type-2 fuzzy labels. The loss function is used to train the parameters of the fuzziness intializer. Finally, the interval type-2 fuzzy labels are defuzzified to binary predictions. All metrics in our experiments are computed based the binary predictions and groundtruth binary labels.}
\label{fig:scheme}
\end{figure*}
\indent However, if we are interested in multi-label classifications, there is a key difference between probabilities and fuzzy logic. We should examine the joint probability (Fig.~\ref{fig:fuzzy2}). Instead, we can explain the outputs of a neural network as type-1 fuzzy memberships. Although we give a reasonable explanation to the outputs, type-1 fuzzy logic is not fuzzy~\cite{tian-tcyb-2021}. Hence, interval type-2 fuzzy logic is used here. We believe that the more categories an instance belongs to, the fuzzier its memberships are. Therefore, we use the number of its categories to control the fuzziness of its memberships (Fig.~\ref{fig:fuzzy2}). By introducing interval type-2 fuzzy logic, we need design the loss function of the classifier to measure the dissimilarities between interval type-2 fuzzy labels and binary labels.\\

\indent  The overall scheme of our model is illustrated in Fig.~\ref{fig:scheme}. The following of this paper is organized as follows. In Section~\ref{sec:relatedworks}, related works are briefly reviewed. The preliminaries are given in Section~\ref{sec:preliminaries}. The formulation of our models is shown in Section~\ref{sec:formulation}. In Section~\ref{sec:experiments}, we report the experimental results of our method. The conclusive remarks are given in Section~\ref{sec:conclusion}.

\section{Related Works}
\label{sec:relatedworks}
An intuitive way for MLC is using classifier chains~\cite{read_classifier_2009} which decomposes a multi-label classification into a series of binary classifications. The subsequent binary classifiers are built on the prediction of preceding ones. Recurrent neural networks (RNN) and convolutional neural networks (CNN) are used in classifier chains~\cite{wang_cnn-rnn_2016}.\\
\indent The MLC can be also transformed to a label ranking problem. Label ranking based methods use $L(L-1)/2$ binary classifiers for pairwise comparisons between each label pair~\cite{calibrate-label-ranking}. Wang~\emph{et al.}~\cite{wang-tcyb-2018} proposed an efficient learning algorithm for solving multi-label ranking problem which is used to model facial action unit recognition task.\\
\indent Random $k$-Labelsets~\cite{random-k-labelsets}\cite{active-k-labelsets} methods build multi-class sub-models on random subsets of labels, and then learn single-label classifier for the prediction of each element in the powerset of this subset. Lo~\emph{et al.}~\cite{generalized-k-labelsets} proposed a basis expansions model for MLC. The basis function is a label powerset classifier trained on a random $k$-labelset. The authors derive an analytic solution to learn the expansion coefficients which minimize the error between the prediction and the groundtruth. Mutual information is incorporated to evaluate the redundancy level and imbalance level of each $k$-labelset~\cite{mutual-k-labelsets}.\\
\indent K-nearest neighbor (KNN), another classical machine learning method, is combined with maximum a posteriori (MAP) principle to determine the label set of an unseen instance~\cite{MLKNN}. Dimension reduction methods can be also used for MLC~\cite{SLEEC}\cite{chen_feature-aware_2012}\cite{hsu_multi-label_2009}. Xu~\emph{et al.}\cite{xu-tcyb-2022} proposed a probabilistic class saliency estimation approach for calculating the projection matrix in linear discriminant analysis for dimension reduction. Decision tree can be built recursively based on a multi-label-entropy based information gain criterion~\cite{multi-label-decision-tree}. A set of support vector machines can be optimized to minimize an empirical ranking loss for MLC~\cite{Rank-SVM}. The correlation labels can be encoded as constrains for MLC~\cite{collective-mlc}.\\
\indent Aligning embeddings of data features and labels in a latent space is a popular trend in MLC. Yeh~\emph{et al.}~\cite{C2AE} use canonical correlation analysis to combine two autoencoders for features and labels to build a label-correlation sensitive loss function. However, the learned latent space is not smooth so small perturbations can lead to totally different decoding results. Similar decoded targets cannot be guaranteed even though the embeddings of features and labels are close in the latent space~\cite{MPVAE}. Bai~\emph{et al.}~\cite{MPVAE} substitute the autoencoders with variational autoencoders so the deterministic latent spaces become probabilistic latent spaces. KL-divergence is used to align the Gaussian latent spaces and the sampling process enforces smoothness. Sundar~\emph{et al.}~\cite{sundar_out--distribution_2020} adopt the $\beta$-VAE for similar purposes. These methods assume a uni-modal Gaussian latent space, which may cause over-regularization and posterior collapse~\cite{dilokthanakul_deep_2017}\cite{wu_multimodal_2018}. Li~\emph{et al.}~\cite{li-tcyb-2018} proposed a deep label-specific feature learning model to bind the label and local visual region in images and they use two variant graph convolutional networks to capture the relationships among labels. Tan~\emph{et al.}~\cite{tan-tcyb-2022} link the manifolds of instance and label spaces, which facilitates using topological relationship of the manifolds in the instance space to guide the manifold construction of the label space.\\
\indent Modeling label correlation is another popular research area in MLC. Structured prediction energy networks (SPENs)~\cite{SPEN}\cite{EESPEN} optimize the sum of local unary potentials and a global potential are trained with a SVM loss. However, the alternating optimization approach suffers from instabilities during training. Tu~\emph{et al.}~\cite{Tu-2020} proposed several strategies to stabilize the joint training of structured prediction of SPENs. The deep value network (DVN)~\cite{DVN} trains a similar energy network by fitting the task cost function. Zhang~\emph{et al.}~\cite{zhang-tcyb-2020} proposed cross-coupling aggregation strategy to simultaneously exploit the label correlation and handle class-imbalance issue. Ma~\emph{et al.}~\cite{ma-tcyb-2022} assume the instances can be clustered into different groups so that the label correlations can be modeled within each sub-group. \\
\indent Input convex neural networks (ICCNs)~\cite{ICCN} design potentials which are convex with respect to the labels so that inference optimization will be able to reach global optimum. Bi~\emph{et al.}~\cite{bi_multilabel_2014} use a probabilistic model exploiting the multi-label correlations. Lanchantin~\emph{et al.}~\cite{LaMP} propose label message passing (LaMP) neural networks for MLC. LaMP treats labels as nodes on a label-interaction graph and computes the hidden representation of each label node conditioned on the input using attention-based neural message passing. Chen~\emph{et al.}~\cite{chen_multi-label_2019} adopt graph convolutional network (GCN). Each node on the graph is represented by word embeddings of a label. The GCN is learned to map this label graph into a set of inter-dependent object classifiers. These classifiers are applied to the image descriptors extracted by another sub-net, enabling the whole network to be end-to-end trainable. \\
\indent Zhang~\emph{et al.}~\cite{zhang-deep-set-2019} predict labels using a deep neural network in a way that respects the set structure of the problem. Patel~\emph{et al.}~\cite{Box-Embeddings-MLC} exploit the taxonomic relationships among labels using box embeddings~\cite{Box-Embeddings}. Brukhim~\emph{et al.}~\cite{cardinality} use cardinality~\cite{tian-tcyb-early} of labels as constraints for training the classifier.  
\section{Preliminaries}
\label{sec:preliminaries}
We consider the setting of assigning $L$ labels $\mathbf{y}=(y_1,\ldots,y_L)$ to an input $\mathbf{x}\in\mathbb{R}^d$, where $d$ is the dimension of instance features. The true label (which is binary) of $\mathbf{x}$ is denoted as $\mathbf{y}^*$. The fuzziness of $\mathbf{y}$ is controlled by fuzzifier $\underline{m}$ and $\overline{m}$. That is, $\mathbf{x}$ is belonged to the $i$-th category in a degree of $[y_i^{\underline{m}},y_i^{\overline{m}}]$. Almost all datasets provide binary labels. To compute the metrics for comparison, we will generate binary predicted labels $\hat{\mathbf{y}}$ based on $\mathbf{y}$. $\hat{\mathbf{y}}$'s are the final predictions of our model. The notations are listed in Table~\ref{tb:notations}.\\
\begin{table}[h]
\centering
\caption{Notations}
\label{tb:notations}
\begin{tabular}{c | c | l}
\toprule
\emph{notations}&\emph{spaces}&\emph{descriptions}\\
\hline
$\mathbf{x}$&$\mathbb{R}^d$&Instance feature vector\\
$\mathbf{y}$&$[0,1]^L$&Estimated type-1 fuzzy label vector\\
$\mathbf{y}^*$&$\{0,1\}^L$&Groundtruth binary label vector\\
$\bar{\mathbf{y}}$&$\mathbb{R}^L$&Temporary result for defuzzification\\
$\hat{\mathbf{y}}$&$\{0,1\}^L$&Estimated binary label vector\\
$\underline{m}$&$\mathbb{R}^+$&Fuzzifier for lower bound\\
$\overline{m}$&$\mathbb{R}^+$&Fuzzifier for upper bound\\
$d$&$\mathbb{N}^+$&Dimension of feature vectors\\
$L$&$\mathbb{N}^+$&Dimension of label vectors\\
$\alpha,\beta,\gamma$&$\mathbb{R}$&Learned parameters\\
$\eta,\lambda$&$\mathbb{R}^+$&Preset hyperparameters\\
\bottomrule
\end{tabular}
\end{table}

\section{Formulation}
\label{sec:formulation}
In this section, we will explain our models in details. There are three main components in our model. First, there is a deep neural network used as a fuzziness initializer to generate initial labels. Second, we use another network to predict the fuzzifier $\underline{m}$ and $\overline{m}$. Third, we design an algorithm to compare the interval type-2 labels and binary labels.
\subsection{Fuzziness initializer}
Indeed, we can use any classical classification neural networks as our fuzziness initializer. The key difference is the activation functions of output layers. To generate the initial guess of fuzzy membership value, we should use an activation function whose value ranges from 0 to 1. Rather than using Sigmoid function which may cause gradient vanishing and slow convergence, we use the following activation function:
\begin{equation}\label{eq:activation}
\begin{cases} 

g(\mathbf{x})=0.5+\alpha\left(
\mathbf{w}^\top\mathbf{x}-\frac{1}{L}\left({\mathbf{1^\top x+1^\top w}}\right)\right)\\
f(\mathbf{x}) =
\max\left(0,\min\left(1,g(\mathbf{x})\right)\right),
\end{cases}
\end{equation}
where $L$ is the dimension of vector $\mathbf{x}$ or $\mathbf{w}$, and $\alpha$ is a learned parameter which is initially set as 1. $\frac{1}{L}(\mathbf{x+w})$ can be treated as bias term of the neuron. Eq.~\eqref{eq:activation} is similar to the idea of Batch Normalization (BN)~\cite{batch-normalization}. BN learns two parameters $\beta$ and $\gamma$ in
\begin{equation}\label{eq:bn}
\tilde{\mathbf{Z}}=\beta\mathbf{Z}+\gamma,
\end{equation} 
to standardize mini-batches, where $\mathbf{Z}$ is the output of one layer and $\tilde{\mathbf{Z}}$ is the output of BN. $\gamma$ in Eq.~\eqref{eq:bn} is expected to be a constant vector. However, when the neural network is trained by mini-batch, it is impossible for $\gamma$ to hold constant. Hence, for each neuron, we minus the means of $\mathbf{x}$ and $\mathbf{w}$. The bias term of each neuron can vary according to mini-batched data. What Eq.~\eqref{eq:activation} really learns is a proper weight $\mathbf{w}$ that makes $\mathbf{w^\top x}$ can be centered by $\frac{1}{L}\left(\mathbf{x+w}\right)$. $0.5$ in Eq.~\eqref{eq:activation} is used to shift the zero-centered outputs to 0.5-centered so that we can use $f(\mathbf{x})$ to generate values in interval $[0,1]$. $\alpha$ in Eq.~\eqref{eq:activation} is similar to $\beta$ in Eq.~\eqref{eq:bn}. It is used to change the standard deviation of outputs to approximately 1. Therefore, Eq.~\eqref{eq:activation} can avoid gradient vanishing as BN.\\
\subsection{Fuzzifier Estimator} 
Fuzziness estimator is used to estimate the fuzzifier of predicted labels, i.e. $\overline{m}$ and $\underline{m}$. As we discussed above, the fuzziness is related to the number of categories an instance belongs to, i.e. $|y^*|$. Hence, we use another neural network to predict $|y^*|$. We denote the prediction of $|y^*|$ as $\hat{m}$ hereafter. \\
\indent To estimate $\hat{m}$, firstly, we train an autoencoder. Then, we train a constrained linear regression model on the outputs of the bottleneck of the autoencoder to estimate $\hat{m}$ (Fig.~\ref{fig:scheme}). \\
\indent We add a softmax layer on the bottleneck of the autoencoder and we add a cross-entropy to the object function of the autoencoder:
\begin{equation} \label{eq:autoencoder}
\begin{aligned}
\mathop{\arg\min}_{\Theta_e,\Theta_d}\left\|x-h\left(h\left(\mathbf{x};\Theta_e\right);\Theta_d\right)\right\|^2_F\\
+\eta CE\left(S\left(h\left(\mathbf{x};\Theta_e\right),l\right)\right),
\end{aligned}
\end{equation}
where $h(\mathbf{x},\Theta_e)$ is the function of encoder, $h(\mathbf{x},\Theta_d)$ is the function of decoder, $CE$ is cross entropy function, $S$ is the Softmax layer and $\eta$ is a tuned parameter. $l$ is an one-hot vector which indicates the number of categories an instance belongs to. If the $i$-th element of $l$ is non-zero, it indicates that the instance belongs to $i$ categories. For example, if the original label of an instance is $[0,1,0]$, its $l$ is $[1,0,0]$ which indicates the instance only belongs to one category. If the original label of an instance is $[1,0,1]$, its $l$ is $[0,1,0]$ which indicates the instance belongs to two categories.\\
\indent The reason why we do not directly train a classification neural network to predict $l$ is that we want to de-correlate the neural networks we used for predicting $y$ and $l$. This idea comes from Random Forest which randomly chooses subsets of variables to construct trees so that these trees are less correlated. The Random Forest increases the bias of the model while decreases the variance of the model. In our model, although $y$ and $l$ are different supervision information, they are correlated to each other. Furthermore, the neural networks have the ability of predicting $y$ or $l$ using a few layers close to the output, even when the inputs are the same. That is, if we use the same random seed and structure for two deep neural networks that predict $y$ and $l$ respectively, they may only significantly differ in 1 or 2 layers close to the output. Using autoencoder, the data themselves are another ``supervision'' information for training so that those two neural networks are possible to be more de-correlated.\\
\indent After the autoencoder is trained, we use the bottleneck's outputs to train a simple linear regression model. The label is the number of categories an instance belongs to, i.e. $|y^*|$. We use the output of the linear regression model as $\hat{m}$. We set $\underline{m}=\hat{m}/|y^*|$ and $\overline{m}=\hat{m}/L$. Note that $|y^*|\leq L$ and $0\leq y\leq 1$. Therefore, $\underline{m}\geq \overline{m}$ and $y^{\underline{m}}\leq y^{\overline{m}}$. For a special case when $\overline{m}=\underline{m}$, the fuzzifier estimator believes the instance only belongs to one category. There is no fuzziness at all in this case.
\subsection{Loss Function}
Traditional loss functions for structured learning can be cross entropy, $F_1$ measure, etc. Our model generates interval type-2 fuzzy labels for data. However, the real labels for the data are generally binary. Hence, we should make a judgment on the differences between the predicted labels $\mathbf{y}$ and real labels $\mathbf{y}^*$. An intuitive way is using the middle point of each interval $[y_i^{\underline{m}},y_i^{\overline{m}}]$. Then, the traditional loss functions can be applied. The problem is that two different intervals may have the same middle point. For example, $[0.3,0.6]$ and $[0.4,0.5]$ have the same middle point $0.45$. This way cannot sufficiently utilize the fuzziness information. \\
\indent We design our loss function based on the continuous extension of $F_1$ measure~\cite{DVN}:
\begin{equation}\label{eq:loss-function}
E\left(\mathbf{y},\mathbf{y}^*,\underline{m},\overline{m}\right) = -\frac{2{\mathbf{y}^{\underline{m}}}^\top\mathbf{y}^*}{\mathbf{1}^\top\left(\mathbf{y}^{\underline{m}}+\mathbf{y}^*\right)}-\frac{2{\mathbf{y}^{\overline{m}}}^\top\mathbf{y}^*}{\mathbf{1}^\top\left(\mathbf{y}^{\overline{m}}+\mathbf{y}^*\right)}
\end{equation}
\subsection{Defuzzification}
To compute the metrics for comparing different methods. We eventually have to generate binary prediction of labels. That is, we have to determine an instance belongs to which categories based on their fuzzy memberships. Suppose we use the middle point of an interval for defuzzification. Then, we just need to sort the middle points, and set top $[\hat{m}]$ $y_i$'s as 1 and the remaining $y_i$'s as 0, where $[\cdot]$ rounds $\cdot$ to the closest integer. The problem is, as we discussed above, the middle point cannot fully utilize the information of interval type-2 fuzzy logic. Here, we use a linear combination between middle point and interval size for defuzzification, i.e.
\begin{equation}\label{eq:defuzzification}
\bar{y}=\frac{y^{\overline{m}}+y^{\underline{m}}}{2} - \frac{\lambda}{2} \left(y^{\overline{m}}-y^{\underline{m}}\right),
\end{equation}
where $\lambda$ is positive constant which is set as 0.1 in our experiments. Finally, we sort the elements in $\bar{y}$, and set the top $[\hat{m}]$ elements as 1 and the remaining elements as 0 to generate final prediction $\hat{y}$.

\subsection{Implementation Details}
\begin{figure}
\centering
\includegraphics[width=0.9\linewidth]{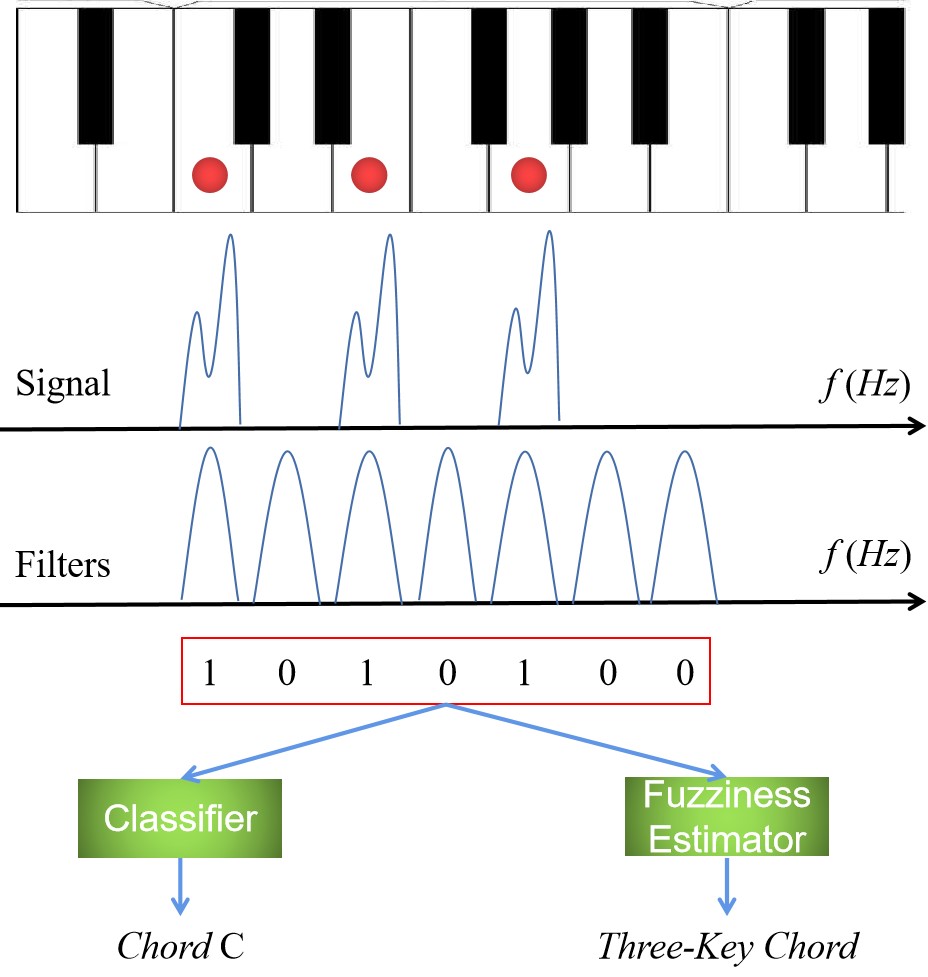}
\caption{Illustration of training fuzzifier estimator. When Chord C is played, the three pressed keys have a spectrum like the second row. If 7 band-pass filters are used to detect which of the 7 white keys in an octave are pressed, the filters' spectrum will look like the third row. Finally, we examine the output of these 7 filters. The filters that have significant outputs are denoted as 1 and others are denoted as 0. We will get the a binary vector at the fourth row. The fuzziness intializer identifies which chord is played, while the fuzzifier estimator only counts how many keys are pressed, which is equivalent to compute the $l_0$-norm of filters' outputs.}
\label{fig:piano}
\end{figure}

Fuzziness initializers can be any classification neural networks as long as their output layers can be written as an activation imposing on a linear combination of weights and inputs. We use $\mathcal{B}=\{b_i\},i=1,\ldots$ to represent the set that contains all such neural networks. $b_i$ is a single candidate neural network that can be used as our fuzziness initializer. We can directly substitute the activation function of the output layer of $b_i$ with Eq.~\eqref{eq:activation}. Otherwise, if we want to directly use the pre-trained weights, we can add an additional layer with activation function Eq.~\eqref{eq:activation} on the top of the output layer of $b_i$. \\
\indent For computational efficiency, we do not have to train the autoencoder separately. We can use several layers of the fuzziness initializer $b_i$ as the encoder, build a symmetric neural network as decoder and freeze all other parameters. Let us take VGG16~\cite{vgg} as an example. VGG16 is a convolutional neural network for image classification. We can freeze all the parameters of convolutional layers. Then we use the full-connected layers as encoder and build a symmetric full-connected layers as decoder. Thus, the input data $\mathbf{x}$ of autoencoder in Eq.~\eqref{eq:autoencoder} is actually the outputs of convolutional part or the inputs of full-connected part of VGG16. The rationale of this is illustrated in Fig.~\ref{fig:piano}. Note that in the view of signal processing, convolution in time domain is multiplication in frequency domain. One can use Fourier Transform to compute the spectrum of a time-domain signal.\\
\indent In Fig~\ref{fig:piano}, both of the fuzziness intializer and the fuzzifier estimator can use the same vector for their own predictions. For convolutional neural networks, such as VGG16, we treat the convolutional part as feature extraction. Fuzziness intializer and fuzzifier estimator share the features. Thus, after we trained the fuzziness intializer, we can freeze the feature extraction part. On the other hand, we can directly use pretrained deep neural networks to extract feature vectors from raw data. In this way, deep full-connected neural networks of several layers can be used both for the fuzziness initializer and the fuzzifier estimator.\\
\section{Experimental Results}
\label{sec:experiments}
\subsection{Benchmarks}
We evaluate our method on three widely-used multi-label datasets, \emph{scene}~\cite{scene}, \emph{mirflickr}~\cite{mirflickr} and \emph{nus-wide}~\cite{nus-wide}. These datasets are publicly available\footnote{\emph{scene} and \emph{nus-wide} datasets with 128D-cVALD features are available at http://mulan.sourceforge.net/datasets-mlc.html.  \emph{mirflickr} dataset is available at https://github.com/JunwenBai/c-gmvae.}. The statistics of these three datasets are given in Table~\ref{tb:datasets}. We randomly separate \emph{nus-wide} dataset into training (80\%), validation (10\%) and tesing (10\%) splits and use the original splits of \emph{scene} and \emph{mirflickr} datasets. To eliminate the effects of feature extraction neural networks, we directly use extracted features rather than raw images. For \emph{scene} and \emph{mirflickr} datasets, there is only one option on features in the download links. For \emph{nus-wide} dataset, we select the 128D-cVALD features.
\begin{table*}
\centering
\caption{Datasets Statistics}
\label{tb:datasets}
\begin{tabular}{c|c c c c c}
\toprule
Dataset&No. Samples&No. Labels&Feature Dimension&Mean Labels Per Sample&Mean Samples Per Label\\
\hline
\emph{scene}&2407&6&294&1.07&170.83\\
\emph{mirflickr}&25000&1000&38&4.8&1247.34\\
nus-vec&269648&128&85&1.86&3721.7\\
\bottomrule
\end{tabular}
\end{table*}
\subsection{Baselines}
We compare our methods to ASL~\cite{ASL}, RBCC~\cite{RBCC}, MPVAE~\cite{MPVAE}, LaMP~\cite{LaMP}, C2AE~\cite{C2AE}, SLEEC~\cite{SLEEC}, HARAM~\cite{HARAM}, MLKNN~\cite{MLKNN} and BR~\cite{BR}.\\
\indent ASL uses an asymmetric loss for positive and negative samples so that easy negative samples can be dynamically down-weighted. RBCC learns a Bayesian network to condition the prediction of child nodes only on their parents in order to circumvent the unprincipled ordering in classical recurrent classifier chains~\cite{RCC}. MPVAE learns and aligns probabilistic embedding for labels and features. LaMP uses attention-based neural message passing to compute the hidden representation of label nodes of a label-interaction graph to learn the interactions among labels. C2AE adopts autoencoders to embedding data and labels in a latent space. SLEEC assumes the label matrix is low-rank so embedding the label in low-dimensional space can improve the prediction accuracy. HARAM adds an extra ART layer to fuzzy adaptive resonance associative map neural network to increase the classification speed. MLKNN first finds the k-nearest neighbors of an instance and then uses maximum a posteriori principle to determine the label set. BR decomposes the multi-label classification into independent binary classifications. 
\subsection{Evaluation}
We evaluate our method using four metrics, i.e. Example F1, Micro-F1, Macro-F1 and Hamming Accuracy (HA). We denote true positives, false positives and false negatives by $TP_j$, $FP_j$ and $FN_j$, respectively for the $j$-th of $L$ label categories. HA is defined as 
\begin{equation}
\frac{1}{L}\sum_{j=1}^L\mathbbm{1}\left[\hat{y}_j=y^*_j\right],
\end{equation}
where $\mathbbm{1}[\cdot]$ is an indicator function which equals to 1 if the condition is true otherwise equals to 0. Example-F1 is defined as
\begin{equation}
\frac{2\sum_{j=1}^L\hat{y}_jy^*_j}{\sum_{j=1}^Ly_j+y^*j}.
\end{equation}
Micro-F1 is defined as
\begin{equation}
\frac{\sum_{j=1}^Ltp_j}{\sum_{j=1}^L2tp_j+fp_j+fn_j}.
\end{equation}
Macro-F1 is defined as
\begin{equation}
\frac{1}{L}\sum_{j=1}^L\frac{2tp_j}{2tp_j+fp_j+fn_j}.
\end{equation}
\subsection{Results and discussion}
\begin{table}
\centering
\tabcolsep=0.1cm
\caption{example-F1 and micro-F1 scores of different methods on all datasets}
\label{tb:F1-1}
\begin{tabular}{c|c c c||c c c}
\toprule
Metric&\multicolumn{3}{c||}{example-F1}&\multicolumn{3}{c}{micro-F1}\\
\hline
Dataset&\emph{scene}&\emph{mirflickr}&\emph{nus-wide}&\emph{scene}&\emph{mirflickr}&\emph{nus-wide}\\
\hline
BR&0.606&0.325&0.343&0.706&0.371&0.371\\
MLKNN&0.691&0.383&0.342&0.667&0.415&0.368\\
HARAM&0.717&0.432&0.396&0.693&0.447&0.415\\
SLEEC&0.718&0.416&0.431&0.699&0.413&0.428\\
C2AE&0.698&0.501&0.435&0.713&0.545&0.472\\
LaMP&0.728&0.492&0.376&0.716&0.535&0.472\\
MPVAE&0.751&0.514&0.468&0.742&0.552&0.492\\
ASL&0.770&0.477&0.468&0.753&0.525&0.495\\
RBCC&0.758&0.468&0.466&0.749&0.513&0.490\\
\hline
\textbf{Ours}&\textbf{0.782}&\textbf{0.536}&\textbf{0.483}&\textbf{0.771}&\textbf{0.582}&\textbf{0.521}\\
\bottomrule
\end{tabular}
\end{table}
\begin{table}
\centering
\tabcolsep=0.1cm
\caption{macro-F1 and HA scores of different methods on all datasets}
\label{tb:F1-2}
\begin{tabular}{c|c c c||c c c}
\toprule
Metric&\multicolumn{3}{c||}{macro-F1}&\multicolumn{3}{c}{HA}\\
\hline
Dataset&\emph{scene}&\emph{mirflickr}&\emph{nus-wide}&\emph{scene}&\emph{mirflickr}&\emph{nus-wide}\\
\hline
BR&0.704&0.182&0.083&0.901&0.886&0.971\\
MLKNN&0.693&0.266&0.086&0.863&0.877&0.971\\
HARAM&0.713&0.284&0.157&0.902&0.634&0.971\\
SLEEC&0.699&0.364&0.135&0.894&0.870&0.971\\
C2AE&0.728&0.393&0.174&0.893&0.897&0.973\\
LaMP&0.745&0.387&0.203&0.903&0.897&0.980\\
MPVAE&0.750&0.422&0.211&0.909&0.898&0.980\\
ASL&0.765&0.410&0.208&0.912&0.893&0.975\\
RBCC&0.753&0.409&0.202&0.904&0.888&0.975\\
\hline
\textbf{Ours}&\textbf{0.772}&\textbf{0.445}&\textbf{0.231}&\textbf{0.917}&\textbf{0.908}&\textbf{0.986}\\
\bottomrule
\end{tabular}
\end{table}
The results are given in Table~\ref{tb:F1-1} and Table~\ref{tb:F1-2}. Our method outperforms the existing state-of-the-art methods on all datasets. The best numbers are marked in bold. All the numbers of our method are averaged over 5 random seeds. For example-F1, the best results of compared methods are among MPVAE, ASL and RBCC. Our method improves over the best compared method by 1.6\%, 4.3\% and 3.2\% on \emph{scene}, \emph{mirflickr} and \emph{nus-wide}, respectively. For micro-F1 and macro-F1, the best results of compared methods are between MPVAE and ASL. For micro-F1, our method improves over the best compared one by 2.4\%, 5.4\% and 5.3\% on \emph{scene}, \emph{mirflickr} and \emph{nus-wide}, respectively. For macro-F1, our method improves over the best compared one by 0.9\%, 5.5\% and 9.5\% on \emph{scene}, \emph{mirflickr} and \emph{nus-wide}, respectively. For HA, the best results of compared methods are among LaMP, MPVAE and ASL, our method improves over the best one by 0.5\%, 1.1\% and 0.6\% on \emph{scene}, \emph{mirflickr} and \emph{nus-wide}, respectively.
\begin{figure*}
\centering
\includegraphics[width=0.9\linewidth, trim=100 0 100 0]{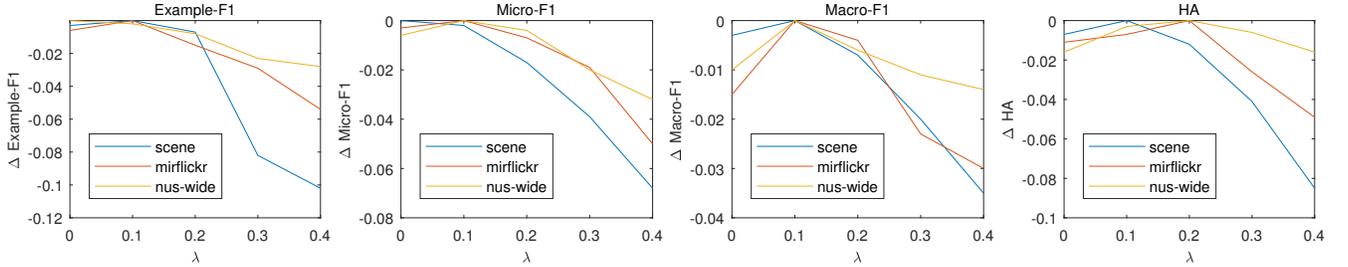}
\caption{$\Delta M$ on different $\lambda$ settings.}
\label{fig:lambda}
\end{figure*}
\subsection{Ablation studies}
To demonstrate the effectiveness of interval type-2 fuzzy logic used in our method, we modify our method by type-1 fuzzy logic. Eq.~\eqref{eq:loss-function} is modified as
\begin{equation}
E(y,y^*)=-\frac{2y^\top y^*}{\mathbf{1}^\top(y+y^*)}.
\end{equation}
For defuzzification, we directly use $\mathbbm{1}[y_j\geq 1/\hat{L}]$. The results on three datasets are given in Table~\ref{tb:ablation}. In Table~\ref{tb:ablation}, example-F1, micro-F1 and macro-F1 are abbreviated as ex-F1, mi-F1 and ma-F1, respectively. The interval Type-2 fuzzy logic substantially improves the performance over type-1 fuzzy logic. \\
\begin{table}
\centering
\caption{Comparison between Interval Type-2 fuzzy logic and type-1 fuzzy logic of our method}
\label{tb:ablation}
\begin{tabular}{c|c|c c c c}
\toprule
&method&ex-F1&mi-F1&ma-F1&HA\\
\multirow{2}{*}{\emph{scene}}&type-1&0.755&0.749&0.753&0.901\\
&type-2&0.782&0.771&0.772&0.917\\
\hline
\multirow{2}{*}{\emph{mirflickr}}&type-1&0.511&0.554&0.412&0.874\\
&type-2&0.536&0.582&0.445&0.908\\
\hline
\multirow{2}{*}{\emph{nus-wide}}&type-1&0.454&0.499&0.208&0.973\\
&type-2&0.483&0.521&0.231&0.986\\
\bottomrule
\end{tabular} 
\end{table}

\indent $\eta$ in Eq.~\eqref{eq:autoencoder} and $\lambda$ in Eq.~\eqref{eq:defuzzification} are two important hyperparameters in our model. $\eta$ controls the weight of cross entropy. In Eq.~\eqref{eq:autoencoder}, both of mean squared error and cross entropy can be independently used as an objective function for training a neural network. We set $\eta$ as 1 in all our experiments. Our model is robust to $\eta$. However, with large $\eta$, one should be careful with the choice of learning rate. Our model is relatively sensitive to $\lambda$. In Fig.~\ref{fig:lambda}, we show the performance of our model in different $\lambda$ settings. In Fig.~\ref{fig:lambda},
\begin{equation}
\Delta M = M_{best} - M_{current}.
\end{equation}
$M$ stands for a metric, i.e. exmaple-F1, micro-F1, macro-F1 or HA. $M_{best}$ stands for the best metric among all experiments of different $\lambda$ settings. $M_{current}$ stands for metric corresponding to current $\lambda$ setting. Hence, all the values of $\Delta M$ are no greater than 0.\\
\indent From Fig.~\ref{fig:lambda}, we found the best $M$'s are generally achieved at $\lambda=0.1\sim 0.2$ and occasionally at $\lambda=0$. When $\lambda=0$, Eq.~\eqref{eq:defuzzification} degrades to classical defuzzification, i.e. the middle point of the interval. When $\lambda$ gets larger, say exceeding $0.3$, the performance drops dramatically. This phenomenon is reasonable. The size of the interval should not dominate the prediction. For example, given two intervals $[0.4,0.5]$ and $[0.5,0.9]$, although the latter interval has a larger size, it still implies the classifier is more confident on the prediction, because all the numbers in the latter interval is no less than those in the former one.
\section{Conclusion}
\label{sec:conclusion}
In this paper, we proposed a multi-label classification method using interval type-2 fuzzy logic. Two neural networks are used for fuzziness intializer and fuzzifier estimation. The fuzzifier estimation neural network is supervised by the number of categories an instance belongs to and its prediction is used to generate interval type-2 fuzzy labels. A loss function measuring the dissimilarities between interval type-2 fuzzy labels and binary labels were proposed for training the fuzziness intializer. Finally, a defuzzification method is proposed to generate binary labels from interval type-2 fuzzy labels for metric computations. Experiments on widely-used datasets validate the better performance of our methods to the compared state-of-the-art methods.\\
\indent Future works include modeling hierarchical relations among labels using interval type-2 fuzzy logic and extending the proposed method to multi-modal multi-label classification task.


\ifCLASSOPTIONcaptionsoff
  \newpage
\fi

\bibliographystyle{IEEEtranS}
\bibliography{references}

\begin{thebibliography}{10}
\providecommand{\url}[1]{#1}
\csname url@samestyle\endcsname
\providecommand{\newblock}{\relax}
\providecommand{\bibinfo}[2]{#2}
\providecommand{\BIBentrySTDinterwordspacing}{\spaceskip=0pt\relax}
\providecommand{\BIBentryALTinterwordstretchfactor}{4}
\providecommand{\BIBentryALTinterwordspacing}{\spaceskip=\fontdimen2\font plus
\BIBentryALTinterwordstretchfactor\fontdimen3\font minus
  \fontdimen4\font\relax}
\providecommand{\BIBforeignlanguage}[2]{{%
\expandafter\ifx\csname l@#1\endcsname\relax
\typeout{** WARNING: IEEEtranS.bst: No hyphenation pattern has been}%
\typeout{** loaded for the language `#1'. Using the pattern for}%
\typeout{** the default language instead.}%
\else
\language=\csname l@#1\endcsname
\fi
#2}}
\providecommand{\BIBdecl}{\relax}
\BIBdecl

\bibitem{ICCN}
B.~Amos, L.~Xu, and J.~Z. Kolter, ``Input convex neural networks,'' in
  \emph{Proceedings of the 34th International Conference on Machine Learning},
  Sydney NSW Australia, 06--11 Aug 2017, pp. 146--155.

\bibitem{MPVAE}
J.~Bai, S.~Kong, and C.~Gomes, ``Disentangled {Variational} {Autoencoder} based
  {Multi}-{Label} {Classification} with {Covariance}-{Aware} {Multivariate}
  {Probit} {Model},'' in \emph{Proceedings of the {Twenty}-{Ninth}
  {International} {Joint} {Conference} on {Artificial} {Intelligence}},
  Yokohama, Japan, Jul. 2020, pp. 4313--4321.

\bibitem{SPEN}
D.~Belanger and A.~McCallum, ``Structured prediction energy networks,'' Jun.
  2016, arXiv:1511.06350.

\bibitem{EESPEN}
D.~Belanger, B.~Yang, and A.~McCallum, ``End-to-end learning for structured
  prediction energy networks,'' in \emph{Proceedings of the 34th International
  Conference on Machine Learning - Volume 70}, Sydney, NSW, Australia, 2017, p.
  429–439.

\bibitem{HARAM}
F.~Benites and E.~Sapozhnikova, ``{HARAM}: {A} {Hierarchical} {ARAM} {Neural}
  {Network} for {Large}-{Scale} {Text} {Classification},'' in \emph{2015 {IEEE}
  {International} {Conference} on {Data} {Mining} {Workshop} ({ICDMW})},
  Atlantic City, NJ, USA, Nov. 2015, pp. 847--854.

\bibitem{SLEEC}
K.~Bhatia, H.~Jain, P.~Kar, M.~Varma, and P.~Jain, ``Sparse local embeddings
  for extreme multi-label classification,'' in \emph{29th {Annual} {Conference}
  on {Neural} {Information} {Processing} {Systems}, {NIPS} 2015, {December} 7,
  2015 - {December} 12, 2015}, Montreal, QC, Canada, December 2015, pp.
  730--738.

\bibitem{bi_multilabel_2014}
W.~Bi and J.~T. Kwok, ``Multilabel classification with label correlations and
  missing labels,'' in \emph{28th {AAAI} {Conference} on {Artificial}
  {Intelligence}, {AAAI} 2014, 26th {Innovative} {Applications} of {Artificial}
  {Intelligence} {Conference}, {IAAI} 2014 and the 5th {Symposium} on
  {Educational} {Advances} in {Artificial} {Intelligence}, {EAAI} 2014, {July}
  27, 2014 - {July} 31, 2014}, Quebec City, QC, Canada, 2014, pp. 1680--1686.

\bibitem{scene}
M.~R. Boutell, J.~Luo, X.~Shen, and C.~M. Brown, ``Learning multi-label scene
  classification,'' \emph{Pattern Recognition}, vol.~37, no.~9, pp. 1757--1771,
  Sep. 2004.

\bibitem{cardinality}
N.~Brukhim and A.~Globerson, ``Predict and constrain: Modeling cardinality in
  deep structured prediction,'' in \emph{Proceedings of the 35th International
  Conference on Machine Learning}, Stockholm Sweden, 10--15 Jul 2018, pp.
  659--667.

\bibitem{chang_2019}
W.-C. Chang, H.-F. Yu, K.~Zhong, Y.~Yang, and I.~Dhillon, ``{X-BERT}: {eXtreme}
  multi-label text classification with using bidirectional encoder
  representations from transformers,'' 2020, arXiv:1905.02331.

\bibitem{chen_feature-aware_2012}
Y.-n. Chen and H.-t. Lin, ``Feature-aware {Label} {Space} {Dimension}
  {Reduction} for {Multi}-label {Classification},'' in \emph{Advances in
  {Neural} {Information} {Processing} {Systems}}, Lake Tahoe Nevada, December
  2012, p. 1529–1537.

\bibitem{chen_multi-label_2019}
Z.-M. Chen, X.-S. Wei, P.~Wang, and Y.~Guo, ``Multi-{Label} {Image}
  {Recognition} with {Graph} {Convolutional} {Networks},'' Apr. 2019,
  arXiv:1904.03582.

\bibitem{nus-wide}
T.-S. Chua, J.~Tang, R.~Hong, H.~Li, Z.~Luo, and Y.~Zheng, ``{NUS}-{WIDE}: a
  real-world web image database from {National} {University} of {Singapore},''
  in \emph{Proceedings of the {ACM} {International} {Conference} on {Image} and
  {Video} {Retrieval}}, New York, NY, USA, Jul. 2009, pp. 1--9.

\bibitem{multi-label-decision-tree}
A.~Clare and R.~D. King, ``Knowledge {Discovery} in {Multi}-label {Phenotype}
  {Data},'' in \emph{Principles of {Data} {Mining} and {Knowledge}
  {Discovery}}, Berlin, Heidelberg, 2001, pp. 42--53.

\bibitem{dilokthanakul_deep_2017}
N.~Dilokthanakul, P.~A.~M. Mediano, M.~Garnelo, M.~C.~H. Lee, H.~Salimbeni,
  K.~Arulkumaran, and M.~Shanahan, ``Deep {Unsupervised} {Clustering} with
  {Gaussian} {Mixture} {Variational} {Autoencoders},'' Jan. 2017,
  arXiv:1611.02648.

\bibitem{Rank-SVM}
A.~Elisseeff and J.~Weston, ``A kernel method for multi-labelled
  classification,'' in \emph{Proceedings of the 14th International Conference
  on Neural Information Processing Systems: Natural and Synthetic}, Cambridge,
  MA, USA, 2001, p. 681–687.

\bibitem{calibrate-label-ranking}
J.~F\"{u}rnkranz, E.~H\"{u}llermeier, E.~Loza~Menc\'{\i}a, and K.~Brinker,
  ``Multilabel classification via calibrated label ranking,'' \emph{Mach.
  Learn.}, vol.~73, no.~2, p. 133–153, nov 2008.

\bibitem{RBCC}
W.~Gerych, T.~Hartvigsen, L.~Buquicchio, E.~Agu, and E.~A. Rundensteiner,
  ``Recurrent {Bayesian} {Classifier} {Chains} for {Exact} {Multi}-{Label}
  {Classification},'' in \emph{Advances in {Neural} {Information} {Processing}
  {Systems}}, 2021, pp. 15\,981--15\,992.

\bibitem{collective-mlc}
N.~Ghamrawi and A.~McCallum, ``Collective multi-label classification,'' in
  \emph{Proceedings of the 14th ACM International Conference on Information and
  Knowledge Management}, New York, NY, USA, 2005, p. 195–200.

\bibitem{DVN}
M.~Gygli, M.~Norouzi, and A.~Angelova, ``Deep value networks learn to evaluate
  and iteratively refine structured outputs,'' in \emph{Proceedings of the 34th
  International Conference on Machine Learning - Volume 70}, Sydney, NSW,
  Australia, 2017, p. 1341–1351.

\bibitem{hsu_multi-label_2009}
D.~Hsu, S.~M. Kakade, J.~Langford, and T.~Zhang, ``Multi-{Label} {Prediction}
  via {Compressed} {Sensing},'' Jun. 2009, arXiv:0902.1284.

\bibitem{mirflickr}
M.~J. Huiskes and M.~S. Lew, ``The {MIR} {Flickr} retrieval evaluation,'' in
  \emph{1st {International} {ACM} {Conference} on {Multimedia} {Information}
  {Retrieval}, {MIR2008}, {Co}-located with the 2008 {ACM} {International}
  {Conference} on {Multimedia}, {MM}'08, {August} 30, 2008 - {August} 31,
  2008}, Vancouver, BC, Canada, 2008, pp. 39--43.

\bibitem{batch-normalization}
S.~Ioffe and C.~Szegedy, ``Batch normalization: Accelerating deep network
  training by reducing internal covariate shift,'' in \emph{Proceedings of the
  32nd International Conference on International Conference on Machine Learning
  - Volume 37}, Lille, France, 2015, p. 448–456.

\bibitem{LaMP}
J.~Lanchantin, A.~Sekhon, and Y.~Qi, ``Neural {Message} {Passing} for
  {Multi}-{Label} {Classification},'' Apr. 2019, arXiv:1904.08049.

\bibitem{li-tcyb-2018}
J.~Li, C.~Zhang, J.~T. Zhou, H.~Fu, S.~Xia, and Q.~Hu, ``Deep-lift: Deep
  label-specific feature learning for image annotation,'' \emph{IEEE
  Transactions on Cybernetics}, vol.~52, no.~8, pp. 7732--7741, 2022.

\bibitem{generalized-k-labelsets}
H.-Y. Lo, S.-D. Lin, and H.-M. Wang, ``Generalized k-labelsets ensemble for
  multi-label and cost-sensitive classification,'' \emph{IEEE Transactions on
  Knowledge and Data Engineering}, vol.~26, no.~7, pp. 1679--1691, 2014.

\bibitem{ma-tcyb-2022}
J.~Ma, B.~C.~Y. Chiu, and T.~W.~S. Chow, ``Multilabel classification with
  group-based mapping: A framework with local feature selection and local label
  correlation,'' \emph{IEEE Transactions on Cybernetics}, vol.~52, no.~6, pp.
  4596--4610, 2022.

\bibitem{RCC}
J.~Nam, E.~L. Menc\'{\i}a, H.~J. Kim, and J.~F\"{u}rnkranz, ``Maximizing subset
  accuracy with recurrent neural networks in multi-label classification,'' in
  \emph{Proceedings of the 31st International Conference on Neural Information
  Processing Systems}, Long Beach, California, USA, 2017, p. 5419–5429.

\bibitem{Box-Embeddings-MLC}
D.~Patel, P.~Dangati, J.~Y. Lee, M.~Boratko, and A.~McCallum, ``Modeling label
  space interactions in multi-label classification using box embeddings,'' in
  \emph{ICLR}, Mar 2022.

\bibitem{read_classifier_2009}
J.~Read, B.~Pfahringer, G.~Holmes, and E.~Frank, ``Classifier {Chains} for
  {Multi}-label {Classification},'' in \emph{Machine {Learning} and {Knowledge}
  {Discovery} in {Databases}}, Berlin, Heidelberg, 2009, pp. 254--269.

\bibitem{ASL}
T.~Ridnik, E.~Ben-Baruch, N.~Zamir, A.~Noy, I.~Friedman, M.~Protter, and
  L.~Zelnik-Manor, ``Asymmetric {Loss} {For} {Multi}-{Label}
  {Classification},'' in \emph{2021 {IEEE}/{CVF} {International} {Conference}
  on {Computer} {Vision} ({ICCV})}, Montreal, QC, Canada, Oct. 2021, pp.
  82--91.

\bibitem{vgg}
K.~Simonyan and A.~Zisserman, ``Very deep convolutional networks for
  large-scale image recognition,'' 2014, arXiv:1409.1556.

\bibitem{sundar_out--distribution_2020}
V.~K. Sundar, S.~Ramakrishna, Z.~Rahiminasab, A.~Easwaran, and A.~Dubey,
  ``Out-of-distribution detection in multi-label datasets using latent space of
  $\beta$-{VAE},'' Mar. 2020, arXiv:2003.08740.

\bibitem{tan-tcyb-2022}
C.~Tan, S.~Chen, G.~Ji, and X.~Geng, ``Multilabel distribution learning based
  on multioutput regression and manifold learning,'' \emph{IEEE Transactions on
  Cybernetics}, vol.~52, no.~6, pp. 5064--5078, 2022.

\bibitem{tian-tcyb-early}
D.~Tian, C.~Gong, M.~Gong, Y.~Wei, and X.~Feng, ``Modeling cardinality in image
  hashing,'' \emph{IEEE Transactions on Cybernetics}, pp. 1--10, Early Access.

\bibitem{tian-tcyb-2021}
D.~Tian, D.~Zhou, M.~Gong, and Y.~Wei, ``Interval type-2 fuzzy logic for
  semisupervised multimodal hashing,'' \emph{IEEE Transactions on Cybernetics},
  vol.~51, no.~7, pp. 3802--3812, 2021.

\bibitem{random-k-labelsets}
G.~Tsoumakas and I.~Vlahavas, ``Random k-{Labelsets}: {An} {Ensemble} {Method}
  for {Multilabel} {Classification},'' in \emph{Machine {Learning}: {ECML}
  2007}, Berlin, Heidelberg, 2007, pp. 406--417.

\bibitem{Tu-2020}
L.~Tu, R.~Y. Pang, and K.~Gimpel, ``Improving joint training of inference
  networks and structured prediction energy networks,'' in \emph{Proceedings of
  the Fourth Workshop on Structured Prediction for NLP}, Online, Nov. 2020, pp.
  62--73.

\bibitem{Box-Embeddings}
L.~Vilnis, X.~Li, S.~Murty, and A.~McCallum, ``Probabilistic embedding of
  knowledge graphs with box lattice measures,'' in \emph{Proceedings of the
  56th Annual Meeting of the Association for Computational Linguistics (Volume
  1: Long Papers)}, Melbourne, Australia, Jul. 2018, pp. 263--272.

\bibitem{wang_cnn-rnn_2016}
J.~Wang, Y.~Yang, J.~Mao, Z.~Huang, C.~Huang, and W.~Xu, ``{CNN}-{RNN}: {A}
  {Unified} {Framework} for {Multi}-label {Image} {Classification},'' Apr.
  2016, arXiv:1604.04573.

\bibitem{mutual-k-labelsets}
R.~Wang, S.~Kwong, Y.~Jia, Z.~Huang, and L.~Wu, ``Mutual information based
  k-labelsets ensemble for multi-label classification,'' in \emph{2018 IEEE
  International Conference on Fuzzy Systems (FUZZ-IEEE)}, Rio de Janeiro,
  Brazil, 2018, pp. 1--7.

\bibitem{active-k-labelsets}
R.~Wang, S.~Kwong, X.~Wang, and Y.~Jia, ``Active k-labelsets ensemble for
  multi-label classification,'' \emph{Pattern Recognition}, vol. 109, no.~C, p.
  107583, Jan 2021.

\bibitem{wang-tcyb-2018}
S.~Wang, G.~Peng, S.~Chen, and Q.~Ji, ``Weakly supervised facial action unit
  recognition with domain knowledge,'' \emph{IEEE Transactions on Cybernetics},
  vol.~48, no.~11, pp. 3265--3276, 2018.

\bibitem{wu_multimodal_2018}
M.~Wu and N.~Goodman, ``Multimodal {Generative} {Models} for {Scalable}
  {Weakly}-{Supervised} {Learning},'' Nov. 2018, arXiv:1802.05335.

\bibitem{xu-tcyb-2022}
L.~Xu, J.~Raitoharju, A.~Iosifidis, and M.~Gabbouj, ``Saliency-based multilabel
  linear discriminant analysis,'' \emph{IEEE Transactions on Cybernetics},
  vol.~52, no.~10, pp. 10\,200--10\,213, 2022.

\bibitem{C2AE}
C.-K. Yeh, W.-C. Wu, W.-J. Ko, and Y.-C.~F. Wang, ``Learning deep latent spaces
  for multi-label classification,'' Jul. 2017, arXiv:1707.00418.

\bibitem{BR}
M.-L. Zhang, Y.-K. Li, X.-Y. Liu, and X.~Geng, ``Binary relevance for
  multi-label learning: an overview,'' \emph{Front. Comput. Sci.}, vol.~12,
  no.~2, pp. 191--202, Apr. 2018.

\bibitem{zhang-tcyb-2020}
M.-L. Zhang, Y.-K. Li, H.~Yang, and X.-Y. Liu, ``Towards class-imbalance aware
  multi-label learning,'' \emph{IEEE Transactions on Cybernetics}, vol.~52,
  no.~6, pp. 4459--4471, 2022.

\bibitem{MLKNN}
M.-L. Zhang and Z.-H. Zhou, ``{ML}-{KNN}: {A} lazy learning approach to
  multi-label learning,'' \emph{Pattern Recognition}, vol.~40, no.~7, pp.
  2038--2048, 2007.

\bibitem{zhang-deep-set-2019}
Y.~Zhang, J.~Hare, and A.~Prügel-Bennett, ``Deep set prediction networks,''
  2019, arXiv:1906.06565.

\end{thebibliography}

\end{document}